\documentclass[10pt,journal,compsoc]{IEEEtran}
\usepackage{balance}
\usepackage{subfigure} 
\usepackage{booktabs}
\usepackage{multirow}
\usepackage{graphics}
\usepackage{epsfig}
\usepackage{mathtools}
\usepackage{amssymb}
\usepackage{makecell}
\usepackage{listings}
\usepackage{algorithm}
\usepackage{color}
\usepackage{algorithmic}
\usepackage{bm}

\begin{document}
\title{PCRDiffusion: Diffusion Probabilistic Models for Point Cloud Registration}

\author{Yue~Wu\IEEEauthorrefmark{1},
	Yongzhe~Yuan\IEEEauthorrefmark{1},
	Xiaolong~Fan,
	Xiaoshui~Huang,
	Maoguo~Gong,~\IEEEmembership{Fellow,~IEEE,} 
	Qiguang~Miao% <-this % stops a space
	\thanks{
		\IEEEauthorrefmark{1} Equal contribution.
	}
\IEEEcompsocitemizethanks{\IEEEcompsocthanksitem Yue~Wu, Yongzhe~Yuan and Qiguang~Miao are with the School of Computer Science and Technology, Key Laboratory of Collaborative Intelligence Systems, Ministry of Education, Xidian University, Xi’an 710071, China.
	E-mail: ywu@xidian.edu.cn, yyz@stu.xidian.edu.cn,   qgmiao@mail.xidian.edu.cn
\IEEEcompsocthanksitem Xiaolong~Fan and Maoguo~Gong are with the School of Electronic Engineering, Key Laboratory of Collaborative Intelligence
	Systems, Ministry of Education, Xidian University, Xi’an 
	710071, China. E-mail: xiaolongfan@outlook.com, gong@ieee.org}% <-this % stops an unwanted space
\thanks{This work is supported by the National Natural Science Foundation of 
	China (62276200, 62036006), the Natural Science Basic Research Plan in 
	Shaanxi Province of China (2022JM-327) and the CAAI-Huawei MINDSPORE 
	Academic Open Fund. (Corresponding author: Maoguo~Gong.)}}


\markboth{Journal of \LaTeX\ Class Files,~Vol.~14, No.~8, August~2015}%
{Shell \MakeLowercase{\textit{et al.}}: Bare Demo of IEEEtran.cls for Computer Society Journals}



\IEEEtitleabstractindextext{%
\begin{abstract}
We propose a new framework that formulates point cloud registration as a denoising diffusion process from noisy transformation to object transformation. During training stage, object transformation diffuses from ground-truth transformation to random distribution, and the model learns to reverse this noising process. In sampling stage, the model refines randomly generated transformation to the output result in a progressive way.
We derive the variational bound in closed form for training and provide implementations of the model. Our work provides the following crucial findings: (i) In contrast to most existing methods, our framework, Diffusion Probabilistic Models for Point Cloud Registration (PCRDiffusion) does not require repeatedly update source point cloud  to refine the predicted transformation.
%, particularly in scenarios where correspondence-free necessitating iterative approaches. 
(ii) Point cloud registration, one of the representative discriminative tasks, can be solved by a generative way and the unified probabilistic formulation. Finally, we discuss and provide an outlook on the application of diffusion model in different scenarios for point cloud registration. Experimental results demonstrate that our model achieves competitive performance in point cloud registration. In correspondence-free and correspondence-based scenarios, PCRDifussion can both achieve exceeding 50\% performance improvements.
\end{abstract}

}

\maketitle

\IEEEdisplaynontitleabstractindextext

% \fi
\IEEEpeerreviewmaketitle

\IEEEraisesectionheading{\section{Introduction}\label{sec:introduction}}

\IEEEPARstart{W}{ith} the rapid development of 3D data acquisition technology \cite{9714804, engel2015large}, point cloud registration, as a fundamental visual recognition task plays an essential role in the 3D vision field, and has been widely applied in various vision tasks, such as 3D scene reconstruction \cite{guo2022neural, siddiqui2021retrievalfuse}, object pose estimation \cite{sun2022onepose, lee2022uda}, and simultaneous localization and mapping (SLAM) \cite{zhu2022nice, fan2022blitz}. Given two 3D point clouds, the goal is to find a rigid transformation to align one point cloud to another. The problem has gained renewed interest recently thanks to the fast growing of 3D point representation learning and differentiable optimization.

\begin{figure}[h]
	\centering
	\includegraphics[width=3.5in]{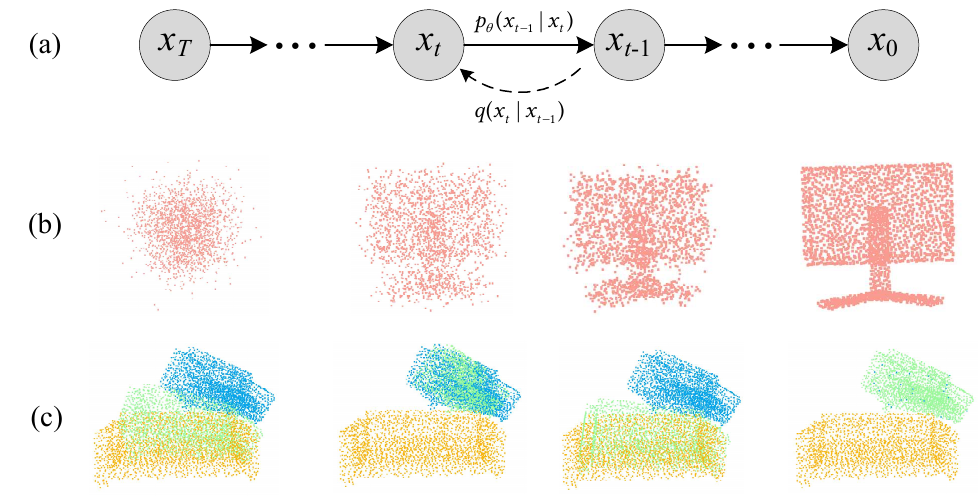}
	\caption{Diffusion model for point cloud registration. (a) A diffusion
		model where $q$ is the forward  process and $p_{\theta}$ is the reverse  process. (b) Diffusion model for point cloud generation task. (c) We propose to formulate point cloud registration as a denoising diffusion process from noisy transfromation to object transfromation. (blue: source point cloud, yellow: template point cloud, green: transformed source point cloud).}
	\label{intro-img1}
\end{figure}

Rigid point cloud registration methods have been evolving in response to the growing complexity of components and usage scenarios. They have advanced from the most basic technique of utilizing regression to predict transformation parameters in the correspondence-free methods \cite{aoki2019pointnetlk, huang2020feature, mei2021point, mei2022partial, xu2021omnet}, employing correspondence-based methods \cite{bai2021pointdsc, choy2020deep, huang2021predator, yew2022regtr} and leveraging SVD decomposition to obtain rigid transformations in partially overlapping scenes. Specifically, in the correspondence-free methods, they are often necessary to seek differences between global features and require them to be sensitive to posture. On the basis of the correspondence-based methods, it is necessary to find the overlapping parts of the point cloud through precise matching and inlier estimation. 
However, point cloud registration methods typically rely on updating source point cloud to further refine the predicted transformation parameters. A natural question is: \textbf{\textit{is there a simpler approach that does not even update source point cloud}?} 

We answer these questions by designing a novel point cloud registration framework that directly predict object transformation from a random transformation. Starting from purely random transformation, which do not contain learnable parameters that need to be optimized in training, we expect to gradually refine the predicted transformation until they perfectly align two point clouds. This strategy does not require to update source point cloud and heuristic object priors, and it has also promoted the development of the point cloud registration pipeline.

Motivated by the discussion above, we propose a new framework that formulates point cloud registration as a denoising diffusion process from noisy transfromation to object transfromation. As shown in Figure \ref{intro-img1}, we argue that the process of transforming noise to the object transformation in point cloud registration is analogous to the process of transforming noise to point clouds in denoising diffusion models. These likelihood-based models gradually remove noise from point clouds to generate point clouds with different shapes. Diffusion models have achieved significant success in various generative tasks \cite{ho2020denoising, austin2021structured, avrahami2022blended, zhou20213d, lyu2021conditional} and have recently started to be explored in some discriminative tasks \cite{amit2021segdiff, baranchuk2021label, brempong2022denoising, chen2022generalist, graikos2022diffusion, kim2022diffusion, wolleb2022diffusion}. However, to the best of our knowledge, there is not existing technique has successfully applied them to point cloud registration.

\begin{figure}[t]
	\centering
	\includegraphics[width=3.6in]{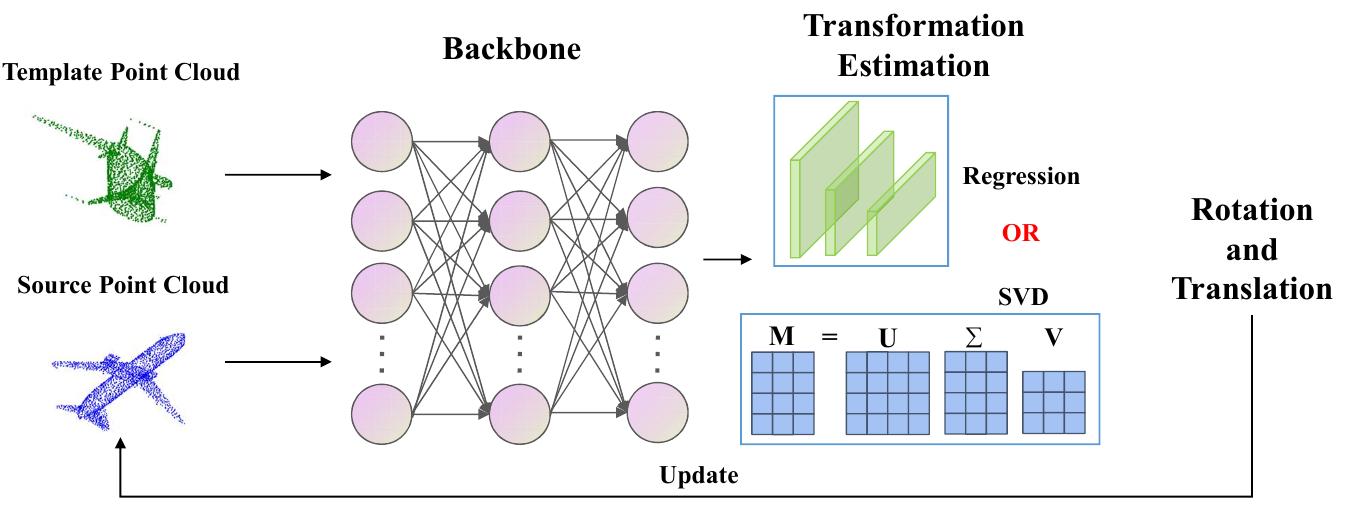}
	\caption{The example of current mature point cloud registration pipeline. Many methods utilize iterative strategy to further refine the predicted transformation parameters.}
	\label{intro-img2}
\end{figure}

In this paper, we propose PCRDiffusion to handle the point cloud registration task with a diffusion model by treating the prediction of transformation parameters as a generation task.
During the training stage, we employ a variance schedule \cite{ho2020denoising, sohl2015deep} to control the addition of Gaussian noise to the ground truth transformation, resulting in a noisy transformation. This noisy transformation is then encoded and utilized to guide the generation of transformations from the output feature of the backbone encoder, such as PointNet \cite{qi2017pointnet}, DGCNN \cite{wang2019dynamic} and PointNet++ \cite{qi2017pointnet++}. Finally, these features are fed into the decoder, which is trained to predict the ground truth transformation without noise. Note that the decoder can be designed based on either correspondence-free methods or correspondence-based methods, depending on the specific context. With this training objective, PCRDiffusion can predict the ground truth transformation from random transformation. During sampling stage, PCRDiffusion generates transformation by reversing the learned diffusion process, which adjusts a noisy prior distribution to the learned distribution over transformation. The ``noise-to-ground truth transformation" pipeline of PCRDiffusion has the appealing advantage of Once-for-All: (i) We train the network only once, and no iteration strategy is required to update source point cloud during training and inference. (ii) We can adjust the number of denoising sampling steps to improve the registration accuracy or accelerate the inference speed. 

To validate the comprehensiveness of the proposed framework, we conduct experiments on both correspondence-based and correspondence-free methods. Specifically, for the correspondence-free method, we employ a fully overlapping experimental setup on the synthetic dataset ModelNet40 \cite{wu20153d}. For the correspondence-based method, we utilize a partially overlapping experimental setup on real-world dataset 3DMatch \cite{zeng20173dmatch} and KITTI \cite{geiger2012we}. Experimental results illustrate that
our framework achieves competitive performance and effectively improves inference speed.
To summarize, our contributions are as follows:
\begin{itemize}
	\item We formulate point cloud registration as a generative denoising
	process, which is the first study to apply the diffusion
	model to point cloud registration to the best of our knowledge.
	\item The "noise-to-ground truth transformation" paradigm has several appealing advantage, such as no iteration strategy is required to update source point cloud and adjust the number of denoising sampling steps to improve the registration accuracy or accelerate the inference speed flexibly.
	\item We conduct experiments on both correspondence-based and correspondence-free methods. Experimental results illustrate that PCRDiffusion achieves competitive performance.
\end{itemize}

The rest of this paper is organized as follows. In Section \ref{sec:related},
we make a  review of the related works of point cloud registration and diffusion models. Section \ref{sec:diffsuion} gives the formulation of the diffusion probabilistic models for point cloud registration, and then introduces PCRDiffsion framework. 
Section \ref{sec:experiment} evaluates the proposed architecture with extensive experiments. Finally, Section \ref{sec:conclusion} concludes this paper and discusses future work.

%-------------------------------------------------------------------
\section{Related Works}
\label{sec:related}
\subsection{Point Cloud Registration}
Most traditional methods need a good initial transformation and
converge to the local minima near the initialization point. One of the most profound methods is the Iterative Closest Point (ICP) algorithm \cite{besl1992method}, which begins with an initial transformation and iteratively alternates between solving two trivial subproblems: finding the closest points as correspondence under current transformation, and computing optimal transformation by SVD \cite{kurobe2020corsnet} based on identified correspondences. Though ICP can complete a high-precision registration, it is susceptible to the initial perturbation and easily prone to local optima. Thus, variants of ICP have been proposed \cite{segal2009generalized, yang2015go, bouaziz2013sparse, fitzgibbon2003robust, rusinkiewicz2019symmetric}, and they can improve the defects of ICP and enhance the registration performance \cite{bellekens2015benchmark}. Generalized-ICP \cite{segal2009generalized} integrates a probabilistic module into an ICP-style framework to enhance the algorithm's robustness.
Go-ICP \cite{yang2015go} combines the ICP that integrated the maximum
correntropy criterion \cite{du2020robust} with the Branch-and-Bound scheme and searches for an optimal solution in the 3D motion space. Besides, Trimmed ICP \cite{chetverikov2002trimmed} introduces the
least trimmed squares into each part of the ICP
pipeline to tackle the partial overlap problem.  
%All these registration methodsstrive to address challenges within the ICP pipeline, yet they fall short of delivering an optimal solution in the absence of a robust initial estimation.
However, all these methods retain a few essential drawbacks. Firstly, they depend strongly on the initialization. Secondly, it is difficult to integrate them into the deep learning pipeline as they lack differentiability. Thirdly, explicit estimation of corresponding points leads to quadratic complexity scaling with the number of points \cite{rusinkiewicz2001efficient}, which can introduce significant computational challenges. 

In order to address the aforementioned problems, learning-based methods have made significant advancements in recent years, which are usually divided into correspondence-based methods \cite{bai2021pointdsc, choy2020deep, huang2021predator, yew2022regtr} and correspondence-free methods \cite{aoki2019pointnetlk, huang2020feature, mei2021point, mei2022partial, xu2021omnet}. Specifically, in the correspondence-free methods, they are often necessary to seek differences between global features and require them to be sensitive to posture. PointNetLK \cite{aoki2019pointnetlk} incorporates the Lucas \& Kanade (LK) algorithm the PointNet \cite{qi2017pointnet} to iteratively
align the input point clouds. In PCRNet \cite{sarode2019pcrnet}, the LK algorithm is substituted with a fully-connected network  and the model improves robustness by regressing.
Correspondence-based methods need to find the correspondences between point clouds through precise matching and inlier estimation. 
IDAM \cite{li2020iterative} introduces a dual stage point elimination method to facilitate the generation of partial correspondences. However, the  efficacy of IDAM remains limited under conditions of low overlapping.
EGST \cite{yuan2023egst} construct reliable geometric structure descriptors to extract correspondences and is less sensitive to outliers. OPRNet \cite{dang2020learning} resorts to the Sinkhorn algorithm \cite{cuturi2013sinkhorn} tailored to partial-to-partial registration. RPMNet \cite{yew2020rpm} utilizes
3D local patch descriptor \cite{deng2018ppfnet} to find correspondences in unorganized point clouds.
However, all methods rely on updating source point cloud to further refine the predicted transformation parameters, as shown in Figure \ref{intro-img2}. In this paper, we aim to push forward the development of the point cloud registration pipeline further with diffusion probabilistic models and  no iteration strategy is required to update source point cloud during training and inference.

\subsection{Diffusion Models}
Diffusion models \cite{ho2020denoising, sohl2015deep, song2019generative, song2020score} have emerged as the cutting-edge family of deep generative models, representing a class of highly advanced deep generative models. They
have broken the long-time dominance of generative adversarial networks (GANs) \cite{goodfellow2020generative} in the challenging task of image synthesis \cite{dhariwal2021diffusion, song2019generative, song2020score} and have also shown potential in computer vision \cite{amit2021segdiff, baranchuk2021label, mathieu2020riemannian}. Specifically, in 3D computer vision, there has been a recent surge of research using generative models for point cloud generation or completion \cite{luo2021diffusion, zhou20213d}. These models are employed to infer missing parts and reconstruct complete shapes and  hold significant implications for various downstream tasks, such as 3D reconstruction, augmented reality, and scene understanding \cite{luo2021score, lyu2021conditional, vahdat2022lion}.
\cite{luo2021diffusion} involves conceptualizing point clouds as particles within a thermodynamic system, utilizing
a heat bath to facilitate diffusion from the original distribution to a noise distribution. The point diffusion-refinement model \cite{lyu2021conditional} is introduced by conditional denoising diffusion probabilistic models to generate a coarse completion from partial observations. Furthermore, this model establishes a point-wise mapping between the generated point cloud and the ground truth.
While diffusion models have achieved great success in generation tasks, their potential for discriminative tasks has yet to be fully explored. Currently, there are some pioneering works that apply diffusion models to image segmentation \cite{wolleb2022diffusion} and object detection \cite{chen2022diffusiondet}. However, despite the considerable interest in this idea, there hasn't been a successful application of diffusion models to point cloud registration solutions before, and its progress lags significantly behind image processing. To the best of our knowledge, this is the first work that adopts a diffusion model for point cloud registration.

%------------------------------------------------------------------------
\section{Diffusion Probabilistic Models for Point Cloud Registration}
\label{sec:diffsuion}
In this section, we begin by introducing the definition of point cloud registration and its mathematical formulations under various modes (Section \ref{sec:pre}). Subsequently, within the context of point cloud registration, we reformulate the diffusion model and establish the training objectives (Section \ref{sec:sf} and Section \ref{sec:rp}). We then provide a detailed exposition of our proposed framework (Section \ref{sec:pcrarch}-Section \ref{sec:sampling}). Finally, we provide the specific details of all experiments (Section \ref{sec:detail}).

\subsection{Preliminaries for Point Cloud Registration}
\label{sec:pre}
We first introduce notations utilized throughout this paper. Given two point clouds: source point cloud $\mathcal{P}=\lbrace{\textbf{p}_i}\in\mathbb{R}^3\mid{i=1,\dots,N}
\rbrace$ and template point cloud $\mathcal{Q}=\lbrace{\textbf{q}_j}\in\mathbb{R}^3\mid{j=1,\dots,M}\rbrace$,
where each point is represented as a vector of $(x, y, z)$ coordinates. 
Point cloud registration task aims to estimate a rigid transformation 
$\lbrace\textbf{R},\textbf{t}\rbrace$ which accurately aligns 
$\mathcal{P}$ and $\mathcal{Q}$, with a 3D rotation 
$\textbf{R}\in SO(3)$ and a 3D translation $\textbf{t}\in \mathbb{R}^3$.
The transformation can be solved by:
\begin{equation}
	\min_{\textbf{R},\textbf{t}}\sum_{(\textbf{p}_i^{*},\textbf{q}_j^{*})\in
		\mathcal{H}^{*}}\|{\textbf{R}\cdot\textbf{p}_i^{*}+
		\textbf{t}-\textbf{q}_j^{*}}\|_2^2,
	\label{eq:corr}
\end{equation}
where $\mathcal{H}^{*}$ is the set of ground-truth correspondences between
$\mathcal{P}$ and $\mathcal{Q}$. 
In this paper, for the convenience of formalization, we merge rotation and translation into a single transformation denoted as \textbf{G}. Note that the meaning of \textbf{G} varies depending on the specific context. In the correspondence-free methods, \textbf{G} represents a 7-dimensional vector composed of a rotation quaternion and a translation. In the correspondence-based methods, \textbf{G} represents a transformation matrix as follow:
\begin{equation}
	\textbf{G}=\begin{bmatrix}
		\textbf{R}_{3\times3} & \textbf{t}_{3\times1}
		\\
		\textbf{0}_{1\times3} & 1
	\end{bmatrix}.
\end{equation}
As a result, our objective becomes:
finding the rigid-body transformation $\textbf{G}\in{SE}(3)$ which best aligns $\mathbf{P}_\mathrm{T}$ 
and $\mathbf{P}_\mathrm{S}$ such that:
\begin{equation}
	\mathcal{Q}=\textbf{G}\cdot\mathcal{P},
\end{equation}
where we use shorthand $(\cdot)$ to denote transformation of $\mathbf{P}_\mathrm{S}$ by rigid 
transform $\textbf{G}$.

\subsection{Formulate Point Cloud Registration as Denoising Diffusion Process}
\label{sec:sf}
The denoising diffusion probabilistic model is a generative model where generation is modeled as a denoising process, in a specific form for point cloud registration:
\begin{equation}
	p_\theta\left(\textbf{G}_0\right):=\int p_\theta\left(\textbf{G}_{0: T}\right) d \textbf{G}_{1: T}.
\end{equation} 
The denoising process begins with standard Gaussian noise and continues until an optimal transformation is achieved. Specifically, a sequence of transformation variables denoted as $\textbf{G}_{T}$, $\textbf{G}_{T-1}$, $\ldots$, $\textbf{G}_{0}$ is generated through the denoising process, where each variable exhibits a progressively reduced level of noise, where $\textbf{G}_{T}$ is sampled from a standard Gaussian prior, and $\textbf{G}_{0}$ represents the final transformation.

We provide a detailed review of the formulation of diffusion models.
Starting from a data distribution $\textbf{G}_0 \sim q\left(\textbf{G}_0\right)$, we define \textit{forward process} $q$ which produces data samples $\textbf{G}_1, \textbf{G}_2, \ldots, \textbf{G}_T$ by gradually adding Gaussian noise at each timestep $t$. In particular, the added noise is scheduled according to a variance schedule $\beta_1, \ldots, \beta_T$:
\begin{equation}
	\begin{aligned}
		q\left(\textbf{G}_{1: T} \mid \textbf{G}_0\right)&:=\prod_{t=1}^T q\left(\textbf{G}_t \mid \textbf{G}_{t-1}\right), \\
		q\left(\textbf{G}_t \mid \textbf{G}_{t-1}\right):=&\mathcal{N}\left(\textbf{G}_t ; \sqrt{1-\beta_t} \textbf{G}_{t-1}, \beta_t \mathbf{I}\right).
	\end{aligned}
\end{equation}
A notable property of the forward process is that it admits sampling $\textbf{G}_t$ at an arbitrary timestep $t$ knowing $\textbf{G}_0$ in convenient closed-form evaluation: using the notation $\alpha_t:=1-\beta_t$ and $\tilde{\alpha}_t:=\prod_{s=1}^t \alpha_s$, we have
\begin{equation}
	\begin{aligned}
		q\left(\textbf{G}_t \mid \textbf{G}_0\right)&:=\mathcal{N}\left(\textbf{G}_t ; \sqrt{\tilde{\alpha}_t} \textbf{G}_0,\left(1-\bar{\alpha}_t\right) \mathbf{I}\right)\\
		&:=\sqrt{\bar{\alpha}_t} \textbf{G}_0+\mathbf{\epsilon} \sqrt{1-\bar{\alpha}_t}, \mathbf{\epsilon} \sim \mathcal{N}(0, \mathbf{I}).
	\end{aligned}
\label{eq:G0t}
\end{equation}
The forward process variances $\beta_t$ can be learned by reparameterization \cite{kingma2013auto}, or held constant as hyperparameters, and expressiveness of the reverse process is ensured in part by the choice of Gaussian conditionals in $p_\theta\left(\textbf{G}_{t-1} \mid \textbf{G}_t\right)$, because both processes have the same functional form when $\beta_t$ are small \cite{sohl2015deep}.  In this work, we
fix the forward process variances $\beta_t$ to constants.

Furthermore, we define \textit{reverse process} $p_\theta(\textbf{G}_{0:T})$ as a Markov chain with learned Gaussian transitions starting at a standard Gaussian prior $p\left(\textbf{G}_T\right)=\mathcal{N}\left(\textbf{G}_T ; \mathbf{0}, \mathbf{I}\right)$:
\begin{equation}
	\begin{aligned}
		p_\theta(\textbf{G}_{0:T})&:=p(\textbf{G}_{T})\prod_{t=1}^{T}p_\theta(\textbf{G}_{t-1}\mid\textbf{G}_{t}),\\
		p_\theta\left(\textbf{G}_{t-1} \mid \textbf{G}_t\right):=&\mathcal{N}\left(\textbf{G}_{t-1} ; \boldsymbol{\mu}_\theta\left(\textbf{G}_t, t\right), \boldsymbol{\Sigma}_\theta\left(\textbf{G}_t, t\right)\right),
	\end{aligned}
\end{equation}
which aims to invert the noise corruption process. Since calculating
$q\left(\textbf{G}_{t-1} \mid \textbf{G}_t\right)$ (ground truth reverse process) exactly should depend on the entire data distribution, we can approximate $q\left(\textbf{G}_{t-1} \mid \textbf{G}_t\right)$ using a neural
network $f_\theta$ with parameter $\theta$, which is optimized to predict a mean $\boldsymbol{\mu}_\theta\left(\textbf{G}_t, t\right)$ and a diagonal covariance matrix $\boldsymbol{\Sigma}_\theta\left(\textbf{G}_t, t\right)$.
Intuitively, the forward process $q\left(\textbf{G}_t \mid \textbf{G}_{t-1}\right)$ can be seen as gradually injecting more random noise to the data, with the reverse process $p_\theta\left(\textbf{G}_{t-1} \mid \textbf{G}_t\right)$ learning to
progressively remove noise to obtain realistic samples by
mimicking the ground truth reverse process $q\left(\textbf{G}_{t-1} \mid \textbf{G}_t\right)$.

\subsection{Reverse Process}
\label{sec:rp}
The goal of training the reverse diffusion process
is to maximize the log-likelihood of the transformation: $\mathbb{E}_{q(\textbf{G}_0)}\left[- \log{p_\theta(\textbf{G}_0)}\right] $. However, since directly optimizing the exact log-likelihood is intractable, we instead maximize its variational lower bound:
\begin{equation}
	\label{eq:training}
	\begin{aligned}
		&\mathbb{E}_{q(\textbf{G}_0)}\left[-\log{p_\theta(\textbf{G}_0)}\right] 
		= -\mathbb{E}_{q(\textbf{G}_0)} \log\left[\int{p_\theta(\textbf{G}_{0:T})} d\textbf{G}_{1:T} \right] \\
		&= -\mathbb{E}_{q(\textbf{G}_0)} \log\left[q(\textbf{G}_{1:T}\mid\textbf{G}_{0}) \frac{p_\theta(\textbf{G}_{0:T})}{q(\textbf{G}_{1:T}\mid\textbf{G}_{0})}d\textbf{G}_{1:T}\right]\\
		&=-\mathbb{E}_{q(\textbf{G}_0)}\log\left[\mathbb{E}_{q(\textbf{G}_{1:T}\mid\textbf{G}_0)}\frac{p_\theta(\textbf{G}_{0:T})}{q(\textbf{G}_{1:T}\mid\textbf{G}_{0})} \right] \\
		&\leq -\mathbb{E}_{q(\textbf{G}_{0:T})}\left[\log\frac{p_\theta(\textbf{G}_{0:T})}{q(\textbf{G}_{1:T}\mid\textbf{G}_{0})} \right]\\
		&= \mathbb{E}_{q(\textbf{G}_{0:T})}\left[-\log\frac{p_\theta(\textbf{G}_{0:T})}{q(\textbf{G}_{1:T}\mid\textbf{G}_{0})} \right]\\
		&=\mathbb{E}_{q(\textbf{G}_{0:T})}\left[-\log p\left(\textbf{G}_T\right)-\sum_{t \geq 1} \log \frac{p_\theta\left(\textbf{G}_{t-1} \mid \textbf{G}_t\right)}{q\left(\textbf{G}_t \mid \textbf{G}_{t-1}\right)}\right] \\
		&=\mathbb{E}_{q(\textbf{G}_{0:T})}\Biggl[-\log p\left(\textbf{G}_T\right)-\sum_{t>1} \log \frac{p_\theta\left(\textbf{G}_{t-1} \mid \textbf{G}_t\right)}{q\left(\textbf{G}_t \mid \textbf{G}_{t-1}\right)}\\
		&\qquad\qquad\qquad\qquad\qquad\qquad\qquad-\log \frac{p_\theta\left(\textbf{G}_0 \mid \textbf{G}_1\right)}{q\left(\textbf{G}_1 \mid \textbf{G}_0\right)}\Biggr]
		\\
		&=\mathbb{E}_{q(\textbf{G}_{0:T})}\Biggl[-\log p\left(\textbf{G}_T\right)-\sum_{t>1} \log \frac{p_\theta\left(\textbf{G}_{t-1} \mid \textbf{G}_t\right)}{q\left(\textbf{G}_{t-1} \mid \textbf{G}_t, \textbf{G}_0\right)} \cdot \\
		&\qquad\qquad\qquad\qquad\qquad\frac{q\left(\textbf{G}_{t-1} \mid \textbf{G}_0\right)}{q\left(\textbf{G}_t \mid \textbf{G}_0\right)}-\log \frac{p_\theta\left(\textbf{G}_0 \mid \textbf{G}_1\right)}{q\left(\textbf{G}_1 \mid \textbf{G}_0\right)}\Biggr] \\
		&=\mathbb{E}_{q(\textbf{G}_{0:T})}\Biggl[-\log \frac{p\left(\textbf{G}_T\right)}{q\left(\textbf{G}_T \mid \textbf{G}_0\right)}-\sum_{t>1} \log \frac{p_\theta\left(\textbf{G}_{t-1} \mid \textbf{G}_t\right)}{q\left(\textbf{G}_{t-1} \mid \textbf{G}_t, \textbf{G}_0\right)}\\
		&\qquad\qquad\qquad\qquad\qquad\qquad\qquad-\log p_\theta\left(\textbf{G}_0 \mid \textbf{G}_1\right)\Biggr]\\
		&=\mathbb{E}_{q(\textbf{G}_{0:T})}\Biggl[\underbrace{D_{\mathrm{KL}}\left(q\left(\textbf{G}_T \mid \textbf{G}_0\right) \| p\left(\textbf{G}_T\right)\right)}_{\mathcal{L}_0}-\underbrace{\log p_\theta\left(\textbf{G}_0 \mid \textbf{G}_1\right)}_{\mathcal{L}_1}\\
		&\qquad\qquad+\sum_{t>1} \underbrace{D_{\mathrm{KL}}\left(q\left(\textbf{G}_{t-1} \mid \textbf{G}_t, \textbf{G}_0\right) \| p_\theta\left(\textbf{G}_{t-1} \mid \textbf{G}_t\right)\right)}_{\mathcal{L}_3}\Biggr],
	\end{aligned}
\end{equation}
where the inequality is by Jensen’s inequality and the above derivation after the fifth row is based on the Bayes' rule \cite{sohl2015deep}.

%The first  term can be ignored as constants since they do not involve any scientific parameters in the forward process, and x is a standard Gaussian noise.

In the final objective of Equation \ref{eq:training}, as the forward process is
fixed and $p(\textbf{G}_{T})$ is defined as a Gaussian prior, $\mathcal{L}_1$ does
not affect the learning of $\theta$.
$\mathcal{L}_0$ can be regarded as the reconstruction of the original data, which can be computed by estimating $\mathcal{N}\left(\textbf{G}_0 ; \boldsymbol{\mu}_\theta\left(\textbf{G}_1, 1\right), \mathbf{\Sigma}_\theta\left(\textbf{G}_1, 1\right)\right)$ and constructing a discrete decoder. Therefore, the ultimate optimization objective is $\mathcal{L}_3$.
Based on Bayes’ theorem, it is found that the posterior
$q\left(\textbf{G}_{t-1} \mid \textbf{G}_t, \textbf{G}_0\right)$ is a Gaussian distribution as well:
\begin{equation}
	q\left(\textbf{G}_{t-1} \mid \textbf{G}_t, \textbf{G}_0\right)=\mathcal{N}\left(\textbf{G}_{t-1} ; \tilde{\mu}\left(\textbf{G}_t, \textbf{G}_0\right), \tilde{\beta}_t \mathbf{I}\right), 
\end{equation}
where
\begin{equation}
	\begin{aligned}
		\tilde{\mu}_t\left(\textbf{G}_t, \textbf{G}_0\right):=&\frac{\sqrt{\tilde{\alpha}_{t-1}} \beta_t}{1-\tilde{\alpha}_t} \textbf{G}_0+\frac{\sqrt{\alpha_t}\left(1-\tilde{\alpha}_{t-1}\right)}{1-\tilde{\alpha}_t} \textbf{G}_t,\\
		&\tilde{\beta}_t:=\frac{1-\tilde{\alpha}_{t-1}}{1-\tilde{\alpha}_t} \beta_t.
	\end{aligned}
\end{equation}
The final training objective
can be reduced to maximum likelihood given the complete
data likelihood with joint posterior $q(\textbf{G}_{1:T}\mid\textbf{G}_{0})$:
\begin{equation}
	\max _\theta \mathrm{E}_{\textbf{G}_0 \sim q\left(\textbf{G}_0\right), \textbf{G}_{1:T} \sim q\left(\textbf{G}_{1:T} \mid \textbf{G}_0\right)}\left[\sum_{t=1}^T \log p_\theta\left(\textbf{G}_{t-1} \mid \textbf{G}_t\right)\right].
\end{equation}

How to choice $p_\theta\left(\textbf{G}_{t-1} \mid \textbf{G}_t\right):=\mathcal{N}\left(\textbf{G}_{t-1} ; \boldsymbol{\mu}_\theta\left(\textbf{G}_t, t\right), \boldsymbol{\Sigma}_\theta\left(\textbf{G}_t, t\right)\right)$ is a crucial question. In this paper, we set $\mathbf{\Sigma}_\theta\left(\textbf{G}_t, t\right)=\sigma_t^2 \mathbf{I}$, where $\sigma_t^2=\tilde{\beta}_t$. To represent $\boldsymbol{\mu}_\theta\left(\textbf{G}_t, t\right)$,  $\frac{1}{\sqrt{\alpha_t}}\left(\textbf{G}_t-\frac{\beta_t}{\sqrt{1-\tilde{\alpha}_t}} \boldsymbol{\epsilon}\right)$ must be predicted given $\textbf{G}_t$. Since $\textbf{G}_t$ is available as input to the model $f_\theta$, we may choose the parameterization:
\begin{equation}
	\begin{aligned}
		\boldsymbol{\mu}_\theta\left(\textbf{G}_t, t\right)=\tilde{\boldsymbol{\mu}}_t\left(\textbf{G}_t, \frac{1}{\sqrt{\tilde{\alpha}_t}}\left(\textbf{G}_t-\sqrt{1-\tilde{\alpha}_t} \boldsymbol{\epsilon}_\theta\left(\textbf{G}_t\right)\right)\right)\\
		=\frac{1}{\sqrt{\alpha_t}}\left(\textbf{G}_t-\frac{\beta_t}{\sqrt{1-\tilde{\alpha}_t}} \boldsymbol{\epsilon}_\theta\left(\textbf{G}_t, t\right)\right),
	\end{aligned}
\end{equation}
where $\boldsymbol{\epsilon}_\theta$ is the output of $f_\theta$. 
Finally, the training objective is parameterized:
\begin{equation}
	\left\|\boldsymbol{\epsilon}-\boldsymbol{\epsilon}_\theta\left(\sqrt{\bar{\alpha}_t} \mathbf{x}_0+\sqrt{1-\bar{\alpha}_t} \boldsymbol{\epsilon}, t\right)\right\|^2.
\end{equation}
Based on the aforementioned process, the sampling procedure can be expressed as follows:
\begin{equation}
	\textbf{G}_{t-1}=\frac{1}{\sqrt{\alpha_t}}\left(\textbf{G}_t-\frac{\beta_t}{\sqrt{1-\tilde{\alpha}_t}} \boldsymbol{\epsilon}_\theta\left(\textbf{G}_t, t\right)\right)+\sigma_t \mathbf{z}, \mathbf{z} \sim \mathcal{N}(\mathbf{0}, \mathbf{I}).
\end{equation}
Note that since point cloud registration is not a purely generative task, we make further adjustments during the implementation process based on the aforementioned theory, such as the training objective.

\subsection{PCRDiffusion Architecture}
\label{sec:pcrarch}
To demonstrate the generalization of our proposed framework, we conduct separate investigations on both correspondences-based and correspondences-free methods. While the overall algorithmic flow can be shared between the two methods, there are distinct differences in the implementation details of each component. We propose to separate the whole framework into two parts, point cloud encoder and transformation estiamtion decoder, where the former runs only once to extract a global representation or point-wise representation from the raw input point cloud $\mathcal{P}$ and $\mathcal{Q}$, and encodes the noisy transformation calculated by $q\left(\textbf{G}_t \mid \textbf{G}_0\right)$ in Equation \ref{eq:G0t}.
The latter takes these representations as condition to predict object transformation. Notably, for correspondence-free method, the number of input points $N$ is equal to $M$, whereas for correspondence-based method, $N$ is not necessarily equal to $M$.

\noindent{\textbf{Point Cloud Encoder.}}   Point cloud encoder takes as input the raw point cloud and extracts its global/point-wise features for the following pipeline. We implement PCRDiffusion for correspondences-free methods with  five shared Multi-Layered Perceptrons (MLPs) similar to the PointNet \cite{qi2017pointnet} architecture having size 64, 64, 64, 128, 1024. Moreover, PCRDiffusion for correspondences-based methods is implemented by Dynamic Graph CNN \cite{wang2019dynamic}.

\begin{figure*}[h]
	\centering
	\includegraphics[width=7.3in]{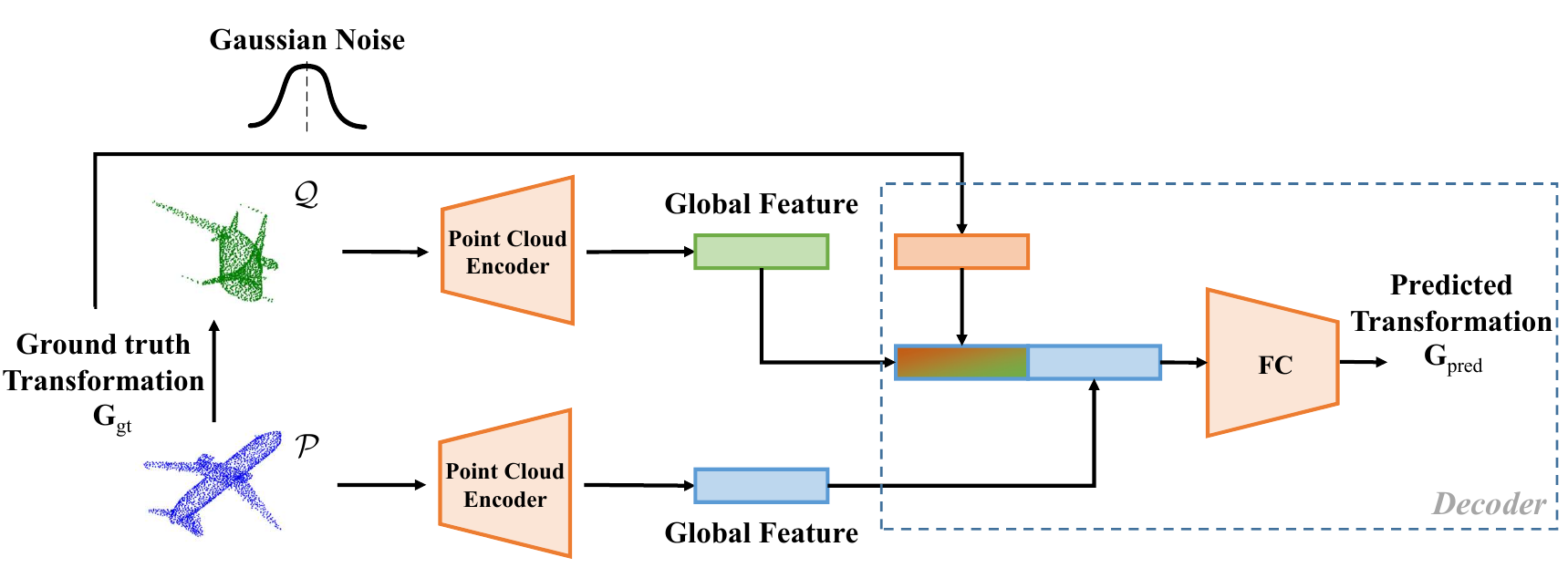}
	\caption{The proposed PCRDiffusion for correspondence-free method during training. Firstly, point cloud encoder  extract global representations from the raw input point cloud $\mathcal{P}$ and $\mathcal{Q}$. Moreover,  the noisy transformation calculated by $q\left(\textbf{G}_t \mid \textbf{G}_0\right)$ in Equation \ref{eq:G0t} is encoded and combined with global representation of $\mathcal{Q}$.
	After that, 
	global representation of $\mathcal{P}$ and combined representation are concatenated and input to FC layers. The predicted transformation  is estimated by difference regression with an 7-dimensional vector (composed of a rotation quaternion and a translation).  }
	\label{meth-corrfree}
\end{figure*}

\begin{figure*}[h]
	\centering
	\includegraphics[width=7.3in]{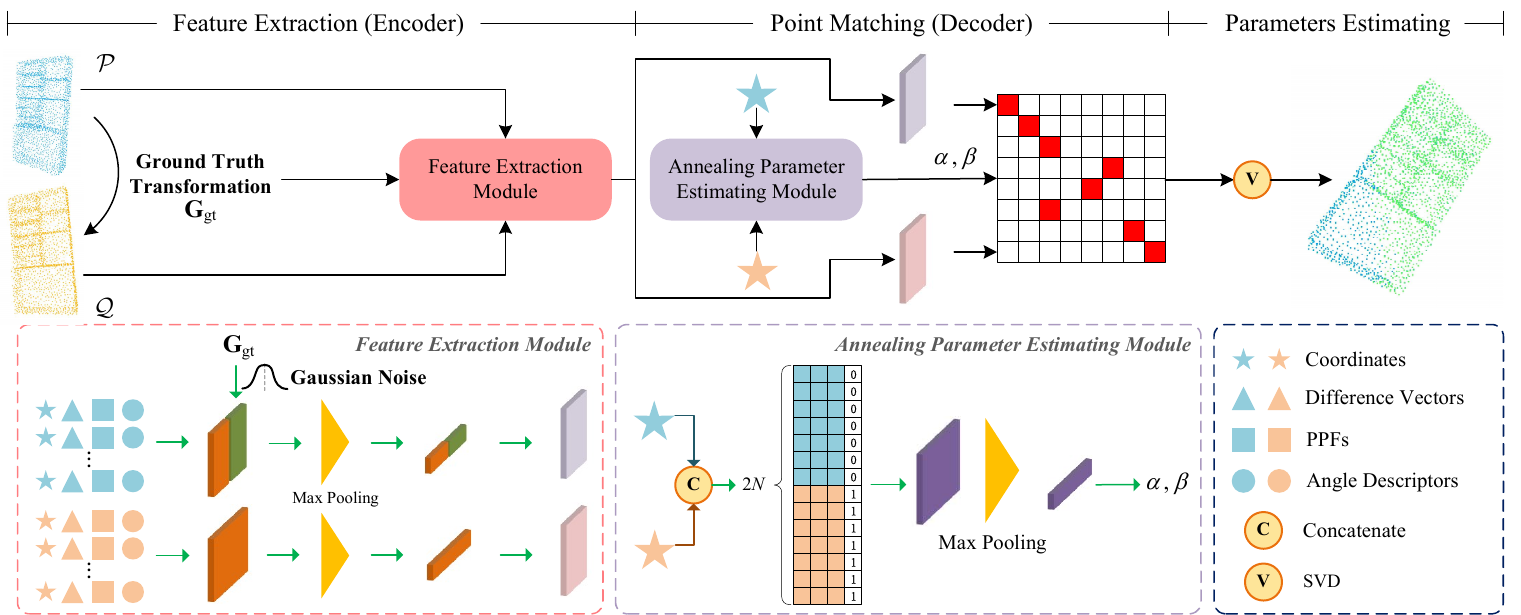}
	\caption{The proposed PCRDiffusion for correspondence-free method during training. We explicitly construct three geometric descriptors based on the original coordinates, which can describe external
		and internal structure of point clouds commendably, and combine them into transformation-invariant structured feature. Feature Extraction Module
		encodes the structured feature and learns contextual features of the geometric structure across
		the two point clouds $\mathcal{P}$ and $\mathcal{Q}$ using encoder of EGST. Moreover, the noisy transformation calculated by $q\left(\textbf{G}_t \mid \textbf{G}_0\right)$ in Equation \ref{eq:G0t} is encoded and combined with point-wise structured feature of $\mathcal{Q}$. Point Matching part use learnable RPM to extract correspondences by Annealing Parameter
		Estimating Module. Finally, the transformation is estimated with SVD method.}
	\label{meth-corrbased}
\end{figure*}

\noindent{\textbf{Transformation Estiamtion Decoder.}} The decoding method varies depending on whether the search for correspondences is required. As shown in Figure \ref{meth-corrfree}, for correspondences-free method, it searches the difference between the global features of source point cloud and template point
cloud, and estimate the rigid transformation by difference
regression with Multi-Layer Perceptrons (MLPs). As shown in Figure \ref{meth-corrbased}, for correspondences-based method, we explicitly construct three geometric
descriptors utilizing different methods, and combine them
into transformation-invariant structured feature.  A feature extraction module is utilized to encode the
structured feature and simultaneously model the structure
consistency across the two point clouds. Based
on previous pipeline, we use RPM which has learnable
parameters to extract correspondences. Finally,
we recover the alignment transformation from the point
matching computed in prior component. We opt EGST \cite{yuan2023egst} to establish our
decoder network, which produces meaningful results in real-world challenges.

\subsection{Training}
\label{sec:train}
During training, we first construct the diffusion process
from ground-truth transformation to noisy transformation and then train the model to reverse this process. Algorithm \ref{alg:train} provides the pseudo code of PCRDiffsuion training procedure.

\noindent \textbf{Ground Truth Transformation Corruption.} We add Gaussian noises to the 
ground truth transformation by $q\left(\textbf{G}_t \mid \textbf{G}_0\right)$. The noise scale is controlled by $\alpha_t$, which adopts the monotonically decreasing cosine
schedule for $\alpha_t$ in different time step $t$, as proposed in \cite{nichol2021improved}. A crucial point to note is that we uniformly transform the initial input ground truth transformation matrix $\textbf{G}_{gt}$ into 7-dimensional vectors (composed of a rotation quaternion and a translation) before encoding, instead of directly utilizing the transformation matrix. This strategy offers several advantages. Firstly, it ensures linearity in the spread of objects while maintaining same significance. Secondly, it reduces the complexity required for the design of the transformation encoder network.

\noindent \textbf{Training Loss.}  We employ a supervised training approach in various modes and have tailored different loss functions for each mode. During training, a neural network
$f_\theta(\textbf{G}_{t}, t)$ is trained to predict $\textbf{G}_{0}$ from $\textbf{G}_{t}$ by minimizing the
training objective with $l_2$ loss :
\begin{equation}
	\mathcal{L}_{diff}=\frac{1}{2}\left\|f_\theta(\textbf{G}_{t}, t)-\textbf{G}_{0} \right\| ^2.
\end{equation}
Note that since point cloud registration is not a purely generative task, we make further adjustments during the implementation process based on the aforementioned theory in Equation \ref{eq:training}. For correspondence-free methods, we optimize PCRDiffusion unioning the following two loss functions:
\begin{equation}
	\mathcal{L}_{cf1}=\frac{1}{N}\sum_{\textbf{p}\in{\mathcal{P}}}\min_{\textbf{q}
		\in\mathcal{Q}}
	\|\textbf{p}-\textbf{q}\|_2^2
	+\frac{1}{M}\sum_{\textbf{q}\in\mathcal{Q}}\min_{\textbf{p}\in{\mathcal{P}}}
	\|\textbf{p}-\textbf{q}\|_2^2,
\end{equation}

\begin{equation}
	\mathcal{L}_{cf2}=\|(\textbf{G}_{0})^{-1}\textbf{G}_{gt}-\textbf{I}_4\|_F.
\end{equation}
For correspondence-based methods, the loss function for joint optimization is identical to that of EGST \cite{yuan2023egst}. It is noteworthy that during the training process, PCRDiffsuion requires only a single pass through the network. Iterative updates by applying the predicted transformation to the source point cloud are unnecessary.

\begin{algorithm}[t]
	\caption{PCRDiffusion Training}
	\label{alg:train}
	%	\algcomment{\fontsize{7.2pt}{0em}\selectfont \texttt{alpha\_cumprod(t)}: cumulative product of $\alpha_i$, \ie, $\prod_{i=1}^t \alpha_i$
		%	}
	\definecolor{codeblue}{rgb}{0.25,0.5,0.5}
	\definecolor{codegreen}{rgb}{0,0.6,0}
	\definecolor{codekw}{RGB}{207,33,46}
	\lstset{
		backgroundcolor=\color{white},
		commentstyle=\fontsize{7.5pt}{7.5pt}\color{codegreen},
		keywordstyle=\fontsize{7.5pt}{7.5pt}\color{codekw},
	}
\begin{lstlisting}[language=python]
def training(ref_pc, src_pc, G_gt):
"""
ref_pc: [B, N, 3]
src_pc: [B, M, 3]
G_gt: [B, 7]
# B: batch size
# N, M: point number
"""

# Encode point cloud 
ref_feats = encoder(ref_pc)
src_feats = encoder(src_pc)

# Corrupt G
# Obtain time step
t = randint(0, T)  

# Noisy transformation: [B, 7]         
eps = normal(mean=0, std=1)  

G_crpt = sqrt(alpha_cumprod(t)) * G_gt + 
         sqrt(1 - alpha_cumprod(t)) * eps

# Encode G_crpt
G_feats = transformation_encoder(G_crpt)

# Predict
G_pred = decoder(G_feats, ref_feats, 
                 src_feats, t)

# Set prediction loss
loss = set_prediction_loss(G_pred, G_gt)

return loss
\end{lstlisting}
\end{algorithm}

\begin{algorithm}[t]
	\caption{PCRDiffusion Sampling}
	\label{alg:sample}
	%	\algcomment{\fontsize{7.2pt}{0em}\selectfont \texttt{alpha\_cumprod(t)}: cumulative product of $\alpha_i$, \ie, $\prod_{i=1}^t \alpha_i$
		%	}
	\definecolor{codeblue}{rgb}{0.25,0.5,0.5}
	\definecolor{codegreen}{rgb}{0,0.6,0}
	\definecolor{codekw}{RGB}{207,33,46}
	\lstset{
		backgroundcolor=\color{white},
		commentstyle=\fontsize{7.5pt}{7.5pt}\color{codegreen},
		keywordstyle=\fontsize{7.5pt}{7.5pt}\color{codekw},
	}
\begin{lstlisting}[language=python]
def sampling(ref_pc, src_pc, T):
"""
ref_pc: [B, N, 3]
src_pc: [B, M, 3]
G_gt: [B, 7]
# B: batch size
# N, M: point number
"""

# Encode point cloud 
ref_feats = encoder(ref_pc)
src_feats = encoder(src_pc)

# Noisy transformation: [B, 7]
G_t = normal(mean=0, std=1)

# Uniform sample step size
times = reversed(linespace(-1, T, steps))

# [(T-1, T-2), ..., (1, 0), (0, -1)]
time_pairs = list(zip(times[:-1], times[1:])

for t_now, t_next in zip(time_pairs):
  # Encode G_t
  G_feats = transformation_encoder(G_t)

  # Predict G_0 from G_t
  G_pred = decoder(G_feats, ref_feats, 
                   src_feats, t_now)

  # Estimate G_t at t_next
  G_t = ddpm_step(G_t, G_pred, t_now, t_next)

return G_pred
	\end{lstlisting}
\end{algorithm}

\subsection{Sampling}
\label{sec:sampling}
The inference procedure of PCRDiffusion is a denoising
sampling process from noise to object transformation $\textbf{G}_{0}$. Starting from
transformation sampled in Gaussian distribution, the model progressively refines its predictions, as shown in Algorithm \ref{alg:sample}. 

\noindent \textbf{Sampling Step.} At sampling stage, data sample $\textbf{G}_{0}$ is reconstructed from
noise $\textbf{G}_{T}$ with the model $f_\theta$ and an updating rule [35, 76] in an iterative way, \textit{i.e.}, $\textbf{G}_{T}$ $\rightarrow \textbf{G}_{T-\Delta} \rightarrow \cdots \rightarrow$  $\textbf{G}_{0}$. In each sampling step, the random transformation or the estimated transformation from the last sampling step are sent
into the transformation estimation decoder. After obtaining the transformation of the current step, DDPM is adopted to estimate the transformation for the next step. 

\noindent \textbf{Once-for-All.}  Thanks to the random transformation design, we can
evaluate PCRDiffusion with an arbitrary number of random
transformation and the number of sampling steps, which do not need
to be equal to the training stage. As a comparison, previous point cloud registration methods typically rely on iterative strategy to further refine the predicted transformation parameters. We train the network only once, and no iteration strategy is required to update source point cloud during training and inference.

\subsection{Implementation Details} 
\label{sec:detail}
In our framework, the learnable parameters are
adjusted by the back propagation algorithm \cite{rumelhart1986learning}. 
We implement and evaluate our model with PyTorch on
an NVIDIA RTX 3090 GPU. The network is trained with
Adam optimizer \cite{kingma2014adam} for 100 epochs on ModelNet40, 50 epochs on 3DMatch and KITTI.  The learning
rate starts from 0.0001 and decays when loss has stopped
improving. The adjustment of the batch size depends on the specific methods used. For correspondence-free method, the batch size is set to 32, while for correspondence-based method, it is set to 8.

\section{Experiments}
\label{sec:experiment}
In this section, we first introduce the experimental settings
and the datasets for point cloud registration. After that, we
compare PCRDiffusion to other diffusion-free methods on the ModelNet40 dataset. Then, we conduct
ablation study to evaluate the effectiveness of the modules
in PCRDiffusion. Lastly, we compare PCRDiffusion to other methods on
the real-world dataset 3DMatch.

\subsection{Setup}
\label{setup}

We evaluate the proposed method on synthetic datasets ModelNet40 \cite{wu20153d}, and real-world dataset 3DMatch \cite{zeng20173dmatch}. ModelNet40 contains 12,308 CAD models of 40 different object categories. Each point cloud 
contains 2,048 points that randomly sampled from the mesh faces and
normalized into a unit sphere. 3DMatch dataset contains 62 indoor subscenes. Some subscenes represent the same scenario with different overlap rates, we ignore this characteristic and treat all subscenes as different scenes. We randomly generate three Euler angle rotations within $[\text{0}^\circ, \text{45}^\circ]$ and translations within $[\text{0}, \text{1}]$ on each axis as the rigid transformation during training.

We compare our method to traditional methods: ICP\cite{besl1992method} and 
FGR\cite{zhou2016fast}, and recent 
learning-based methods: PointNetLK \cite{aoki2019pointnetlk}, PCRNet 
\cite{sarode2019pcrnet}, DeepGMR \cite{yuan2020deepgmr}, 
DCP \cite{wang2019deep}, FMR \cite{huang2020feature} and RPMNet 
\cite{yew2020rpm}. 
We utilize the implementations of ICP and FGR in Intel Open3D 
%\footnote[1]{https://github.com/isl-org/Open3D}
\cite{zhou2018open3d}. 
For other 
baseline, we follow the pipeline provided by original paper. 
For consistency
with previous work, we measure Mean Isotropic Error
(MIE) and Mean Absolute Error (MAE). 
MIE($\textbf{R}$) is the geodesic distance in degrees between estimated and ground-truth rotation matrices. It measures the differences between the predicted and the ground-truth rotation matrices. MIE($\textbf{t}$) is the Euclidean distance between estimated and ground-truth translation vectors. It measures the differences between the predicted and the ground-truth translation vectors.
Here, we give the specific calculation formula of MIE($\textbf{R}$) and MIE($\textbf{t}$):
\begin{equation}
	\begin{aligned}
		\text{MIE}&(\textbf{R})=\text{arccos}\left(\frac{\text{trace}
			(\textbf{R}_{gt}^{-1}\textbf{R}_{est})}{2}\right),\\
		&\text{MIE}(\textbf{t})=\|\textbf{t}_{gt}-\textbf{t}_{est}\|_2.
	\end{aligned}
\end{equation}
Angular measurements are in units of degrees.
All metrics should be zero if reference point cloud align to source point cloud perfectly.

\begin{table*}[h]
	\centering
	\renewcommand\tabcolsep{8pt}
	\caption{Evaluation results on ModelNet40. Bold indicates the best 
		performance and underline indicates the second-best performance.}
	\begin{tabular}{l|ccccccc}
		\toprule[1.5pt]
		%    \multicolumn{2}{c}{Part}                   \\
		%    \cmidrule(r){1-2}
		Method  & \makecell{RMSE($\textbf{R}$)}  & 
		\makecell{RMSE($\textbf{t}$)} & 
		\makecell{MAE($\textbf{R}$)} & \makecell{MAE($\textbf{t}$)} & 
		\makecell{ERROR($\textbf{R}$)}&\makecell{ERROR($\textbf{t}$)}&\# Update $\mathcal{P}$\\
		\midrule
		\multicolumn{8}{c}{(a) Unseen Objects}\\
		\midrule
		%		\multirow{9}{*}{(A)} 
		ICP  \cite{besl1992method} ($\blacktriangle$)&21.2084&0.2874&11.7468&0.1686&23.1548&0.3497&10\\
		PointNetLK  \cite{aoki2019pointnetlk} ($\triangle$)&7.4796&0.5820&1.7362&0.4907&4.0421&0.9741&5\\
		PCRNet  \cite{sarode2019pcrnet}($\triangle$)&19.8457&0.2288&10.2498&0.1168&19.0487&0.2442&8\\
		DeepGMR  \cite{yuan2020deepgmr}($\blacktriangle$)&34.1993&0.4495&23.5162&0.3291&45.7786&0.6758&0\\
		FGR  \cite{zhou2016fast}($\blacktriangle$)&9.5964&0.1186&1.9971&0.0268&4.3337&0.0582&10\\
		DCP  \cite{wang2019deep}($\blacktriangle$)&15.7183&0.2002&10.7846&0.1340&20.8744&0.2740&0\\
		FMR  \cite{huang2020feature}($\triangle$)&6.8525&0.5851&1.5962&0.4886&3.8090&0.9751&5\\
		RPMNet  \cite{yew2020rpm}($\blacktriangle$)&8.0357&0.0841&1.4399&0.0170&2.9619&0.0357&3\\
		\midrule
		PCRDiffusion Correspondence-Free ($\triangle$)&3.3737&0.0460&1.6861&0.0216&3.2786&0.0448  &0\\
		PCRDiffusion Correspondence-Based ($\blacktriangle$)&\textbf{1.0986}&\textbf{0.0218}&\textbf{0.7244}&\textbf{0.0153}&\textbf{1.2732}&\textbf{0.0318}&0\\
		\midrule
		\multicolumn{8}{c}{(b) Unseen Categories}\\
		\midrule
		%		\multirow{9}{*}{(B)} 
		ICP \cite{besl1992method} ($\blacktriangle$)&21.8745&0.2707&11.0308&0.1588&21.9911&0.3307&10\\
		PointNetLK \cite{aoki2019pointnetlk} ($\triangle$)&8.7766&0.5702&2.3269&0.4768&5.5023&0.9555&5\\
		PCRNet \cite{sarode2019pcrnet} ($\triangle$)&22.0931&0.2943&15.4301&0.1946&29.3938&0.3978&8\\
		DeepGMR \cite{yuan2020deepgmr} ($\blacktriangle$)&35.2986&0.4520&22.6519&0.3314&43.7901&0.6816&0\\
		FGR \cite{zhou2016fast} ($\blacktriangle$)&10.6061&0.1214&2.3841&0.0312&5.2151&0.0672&10\\
		DCP \cite{wang2019deep} ($\blacktriangle$)&21.4669&0.3194&15.2743&0.2357&29.3084&0.4834&0\\
		FMR \cite{huang2020feature} ($\triangle$)&7.5793&0.5667&1.8031&0.4750&4.4476&0.9434&5\\
		RPMNet \cite{yew2020rpm} ($\blacktriangle$)&5.3739&0.0736&1.1452&0.0144&2.5072&0.0316&3\\
		\midrule
		PCRDiffusion Correspondence-Free ($\triangle$)&3.1880 &0.0447&1.6887&0.0216&3.1686&0.0446&0\\
		PCRDiffusion Correspondence-Based ($\blacktriangle$)&\textbf{0.2848} & \textbf{0.0045} & \textbf{0.0969}  
		&\textbf{0.0031}&\textbf{0.1790}&\textbf{0.0066}&0\\
		\midrule
		\multicolumn{8}{c}{(c) Gaussian Noise}\\
		\midrule
		ICP \cite{besl1992method} ($\blacktriangle$)&20.3257&0.2696&11.5025&0.1627&22.4326&0.3356&10\\
		PointNetLK \cite{aoki2019pointnetlk} ($\triangle$)&7.5379&0.5771&2.1857&0.4823&4.8129&0.9628&5\\
		PCRNet \cite{sarode2019pcrnet} ($\triangle$)&25.9025&0.3257&14.9771&0.1987&31.5003&0.4115&8\\
		DeepGMR \cite{yuan2020deepgmr} ($\blacktriangle$)&34.0033&0.4573&23.4370&0.3364&45.1482&0.6922&0\\
		FGR \cite{zhou2016fast} ($\blacktriangle$)&12.2387&0.1278&3.1348&0.0387&6.6048&0.0822&10\\
		DCP \cite{wang2019deep} ($\blacktriangle$)&22.2084&0.3312&16.5101&0.2464&31.3406&0.5000&0\\
		FMR \cite{huang2020feature} ($\triangle$)&7.6332&0.5797&2.2957&0.4844&5.1395&0.9651&5\\
		RPMNet \cite{yew2020rpm} ($\blacktriangle$)&9.4977&0.0884&2.7392&0.0310&5.9164&0.0673&3\\
		\midrule
		PCRDiffusion Correspondence-Free ($\triangle$)&3.2549&0.0462&1.6707&0.0220&3.2145&0.0456 &0\\
		PCRDiffusion Correspondence-Based ($\blacktriangle$)&\textbf{1.0766}&\textbf{0.0218}&\textbf{0.7035}&\textbf{0.0152}&\textbf{1.2529}&\textbf{0.0316}&0\\
		\midrule
		\multicolumn{8}{c}{(d) Partial-Partial}\\
		\midrule
		ICP \cite{besl1992method} ($\triangle$)&24.8634&0.3520&14.8698&0.2211&29.1252&0.4610&10\\
		PointNetLK \cite{aoki2019pointnetlk} ($\triangle$)
		&22.1867&0.5812&13.1504&0.4786&26.1196&
		0.9633&5\\
		PCRNet \cite{sarode2019pcrnet} ($\triangle$)
		&21.7402&0.2902&14.2114&0.1990&28.2444&0.4061&8\\
		DeepGMR \cite{yuan2020deepgmr} ($\blacktriangle$)
		&34.6393&0.4706&24.7054&0.3465&47.8755&0.7078&0\\
		FGR \cite{zhou2016fast} ($\blacktriangle$)
		&22.7562&0.2641&8.9966&0.1110&17.2071&0.2289&10\\
		DCP \cite{wang2019deep} ($\blacktriangle$)&24.0230&0.3686&17.8908&0.2772&34.1244&0.5633&0\\
		FMR \cite{huang2020feature} ($\triangle$)
		&15.5280&0.5772&9.5682&0.4777&19.0777&0.9595&5\\
		RPMNet \cite{yew2020rpm} ($\blacktriangle$)&11.0629&0.1912&5.7625&0.1229&11.7768&0.2598&3\\
		%		OMNet \cite{xu2021omnet} ($\triangle$)&4.356&0.0486&1.924&0.0223&3.834&0.0476&4\\
		\midrule
		PCRDiffusion Correspondence-Based ($\blacktriangle$)&\textbf{3.8197} & \textbf{0.0862}& \textbf{2.5673} &\textbf{0.0598}&\textbf{4.7319}&\textbf{0.1250}&0\\
		\bottomrule[1.5pt]
	\end{tabular}
	\label{allresult}
\end{table*}

\subsection{PCRDiffusion for Complete Overlapping Scenarios}
We initially investigate the performance of diffusion models in the complete overlapping scenarios on ModelNet40. 

\begin{figure*}[h]
	\centering
	\includegraphics[width=7.2in]{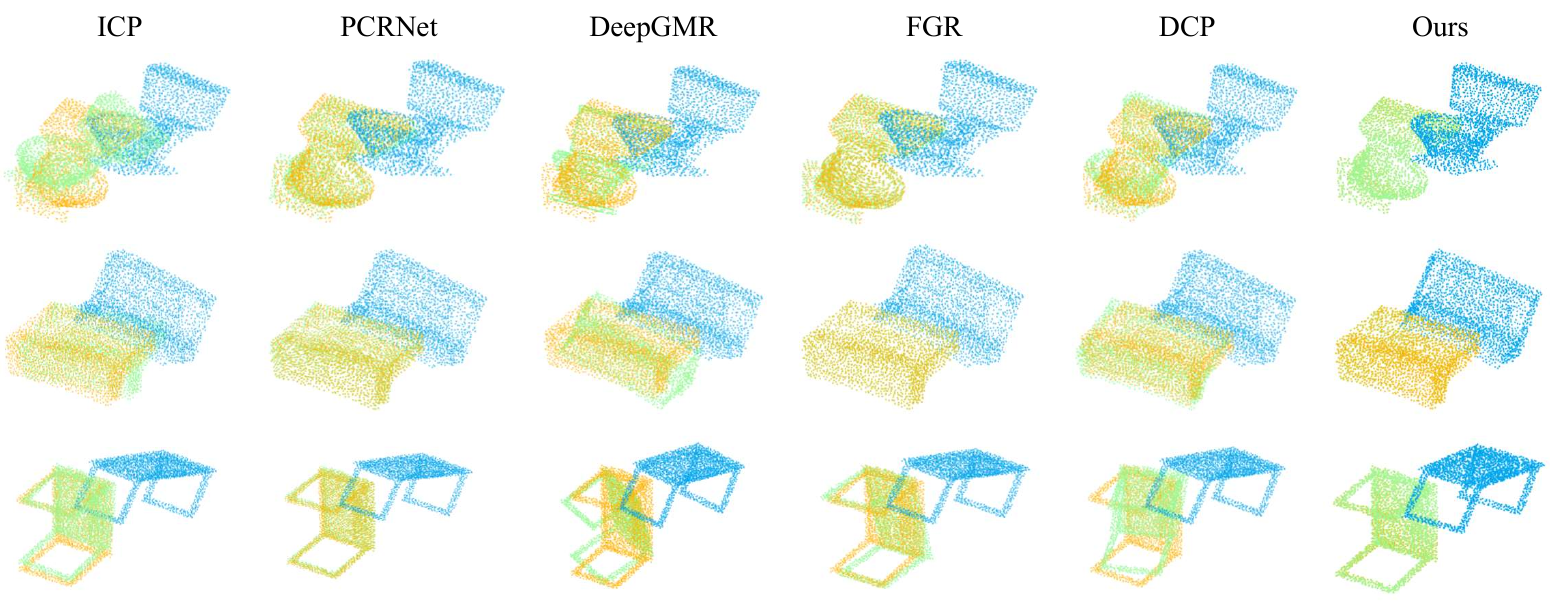}
	\caption{Qualitative comparison of the registration results on unseen objects data (blue: source point cloud, yellow: template point cloud, green: transformed
		source point cloud).}
	\label{fig:corr-free}
\end{figure*}

\noindent \textbf{Unseen Objects.} We train and test the models on the same categories. Training and testing datasets are clean and taken
without any pretreatment. We apply a random transformation
on the template point cloud $\mathcal{P}$ to generate corresponding
source point cloud $\mathcal{Q}$. Table \ref{allresult}(a) shows quantitative results of the various algorithms under current experimental
settings, we observe that PCRDiffusion substantially outperforms
all correspondence-free methods or correspondence-based methods in all metrics. Benefitting
from the diffusion strategy, our
method attains highly robust and accurate registration and
improves the registration accuracy by an order of magnitude. In order to show the effect
of our proposed approach clearly, a qualitative comparison
of the registration results on correspondence-free method can be found in Figure \ref{fig:corr-free}. Our
method ensure minimal impact on changing of the shape.

%\begin{figure*}[t]
%	\centering
%	\fbox{\rule{0pt}{2in} \rule{0.9\linewidth}{0pt}}
%	%\includegraphics[width=0.8\linewidth]{egfigure.eps}
%	\caption{Example of caption.
	%		It is set in Roman so that mathematics (always set in Roman: $B \sin A = A \sin B$) may be included without an ugly clash.}
%	\label{fig:corr-free}
%\end{figure*}

\noindent \textbf{Unseen Categories.}
To verify the robustness of the categories, we evaluate PCRDiffusion by using different categories for training and
testing. ModelNet40 is divided into two parts for training
and testing, each consisting of 20 distinct categories. The
testing portion comprises categories that were not present
during training, thus ensuring that the model is evaluated
on unseen data. Table \ref{allresult}(b) shows the evaluation results of all
models. To evaluate all algorithms fairly, the same categories
and point clouds are used respectively in training and testing.

\noindent \textbf{Gaussian Noise.} 
We evaluate the performance in the presence of noise, which are always
presented in real-world point clouds. 
We apply the random rigid transformation to the template point cloud, and randomly and independently jitter the points in the
source point cloud and template point cloud by noises sampled from
$\mathcal{N}(\text{0}, \text{0.01})$ and clipped to $[-\text{0.05}, \text{0.05}]$ on each axis. 
This experiment is significantly more challenging, because 
finding corresponding points will be more difficult, and the obvious one-to-one correspondence based coordinates will not exist. We retrain all models on the noisy data. As shown in Table \ref{allresult}(c), our method  outperforms other correspondence-free methods and correspondence-based methods.

\subsection{PCRDiffusion for Partially Overlapping Scenarios}

Inspired by Gaussian noise, we observe that our method performs 
well when 
corresponding relationship is found in the cluttered circumstances. Acquisition 
of partial point 
clouds is a common situation in the real-world, therefore we 
challenge a more extensive and difficult task. We crop the template point 
cloud $\mathcal{P}$ and the source point cloud $\mathcal{Q}$ respectively, and retain 
70\% of the points. Noted that, the quantity of points in  two 
point clouds is still consistent after cropping. Moreover, because the 
operation of cropping is random, some points in two point clouds may lose 
their correspondence. We verify quantitatively in Table \ref{allresult}(d) and experimental results 
illustrate that our 
method can 
still achieve stable results comparing other methods, and the other 
baseline do not work well. 

\begin{table*}
	\centering
	\caption{Evaluation results on KITTI and 3DMatch. Bold indicates the best 
		performance.}
	\renewcommand\tabcolsep{3.2pt}
	\renewcommand\arraystretch{1.2}
	\begin{tabular}{l|cccccc|cccccc}
		\toprule[1.5pt]	 
		\multirow{2}{*}{Method} & \multicolumn{6}{c|}{KITTI} 
		&\multicolumn{6}{c}{3DMatch} \\
		&\makecell{RMSE($\textbf{R}$)}  & 
		\makecell{RMSE($\textbf{t}$)} & 
		\makecell{MAE($\textbf{R}$)} & \makecell{MAE($\textbf{t}$)} & 
		\makecell{ERROR($\textbf{R}$)}&\makecell{ERROR($\textbf{t}$)}&
		\makecell{RMSE($\textbf{R}$)}  & 
		\makecell{RMSE($\textbf{t}$)} & 
		\makecell{MAE($\textbf{R}$)} & \makecell{MAE($\textbf{t}$)} & 
		\makecell{ERROR($\textbf{R}$)}&\makecell{ERROR($\textbf{t}$)}\\
		\midrule
		%		\multirow{9}{*}{(A)} 
		ICP \cite{besl1992method}&18.9487&0.5466&12.6160&0.3563&20.6026&0.7752
		&25.4376&0.8635&13.9470&0.4202&25.7505&0.8614\\
		PointNetLK \cite{aoki2019pointnetlk}&7.0309&0.6846&0.8580&0.5257&1.4041
		&1.0589&2.1669&0.5779&0.3797&0.4792&0.6542&0.9653\\
		PCRNet 
		\cite{sarode2019pcrnet}&33.0369&6.8900&22.3317&5.4069&39.5509&10.7703&32.5824&1.3655&25.3004&1.0562
		&49.4726&2.1123\\
		DeepGMR \cite{yuan2020deepgmr}&46.6050&0.4386&25.8622&0.2657&44.0679
		&0.5785&38.3840&1.2166&26.2025&0.8146&48.0517&1.6224\\
		FGR \cite{zhou2016fast}&13.9026&1.3483&1.7946&0.1454&3.2000&0.3053&
		1.5866&0.0810&0.2477&0.0099&0.4764&0.0200\\
		DCP \cite{wang2019deep}&18.1473&10.7517&12.8834&7.4273&22.4225&16.3362&
		29.7272&1.2919&23.0550&0.9689&44.0671&1.9137\\
		FMR 
		\cite{huang2020feature}&7.5326&0.7692&1.4277&0.5663&2.2495&1.1334&3.3293&0.5690&0.6072&
		0.4712&1.0811&0.9423\\
		RPMNet 
		\cite{yew2020rpm}&35.9824&0.7544&19.4641&0.3966&34.4617&0.8745&21.1454&0.4963&8.3353
		&0.2179&15.5514&0.4395\\
		\midrule
		PCRDiffusion &\textbf{0.6281}&\textbf{0.0209}&\textbf{0.3556}&\textbf{0.0129}&\textbf{0.8237}&\textbf{0.0282}&\textbf{0.1522}&\textbf{0.0061}&\textbf{0.1123}&\textbf{0.0040}&\textbf{0.2109}&\textbf{0.0081}
		\\
		\bottomrule[1.5pt]
	\end{tabular}
	\label{realworld}
\end{table*}

\subsection{Other Datasets}
We conduct experiments on the real-world datasets: 3DMatch (indoor) 
\cite{zeng20173dmatch} and KITTI 
odometry (outdoor) \cite{geiger2012we}. We utilize 3D object detection dateset 
in KITTI odometry 
consists of 14,999 outdoor driving scenarios scanned by Velodyne LiDAR and we use sequences 0-5 for training, 6-7 for validation
and 8-10 for testing. 3DMatch 
dataset contains 62 indoor subscenes. Some subscenes represent the same 
scenario with different overlap rates, we ignore this characteristic and treat 
all subscenes as different scenes.  For KITTI, the accuracy here refers to the accuracy of inter-frame matching for a single instance, rather than representing the accuracy of the odometry trajectory after matching all point clouds in the sequence.

\begin{figure*}[h]
	\centering
	\includegraphics[width=7.2in]{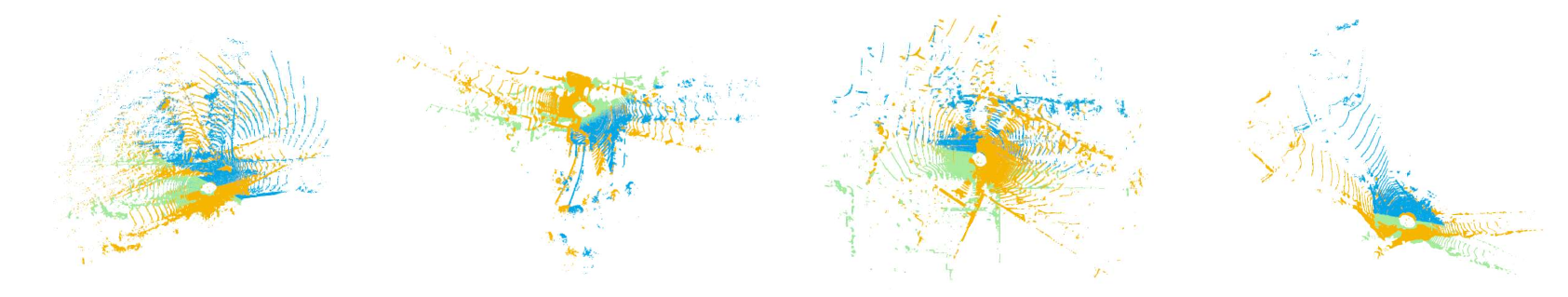}
	\caption{Qualitative results on KITTI odometry dataset(blue: source point cloud, yellow: template point cloud, green: transformed point cloud).}
	\label{kitti-vis}
\end{figure*}

\begin{figure*}[h]
	\centering
	\includegraphics[width=7.2in]{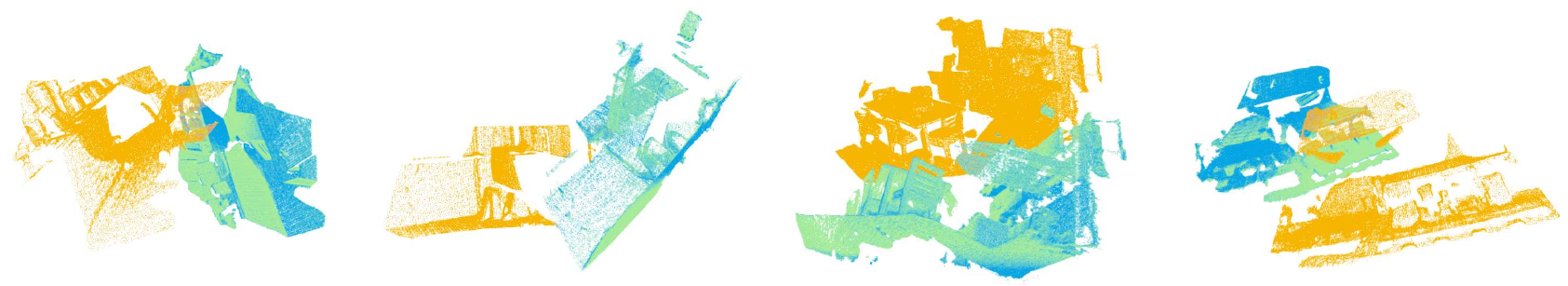}
	\caption{Qualitative results on 3DMatch dataset(blue: template point cloud, yellow: source point cloud, green: transformed point cloud).}
	\label{3dmatch-vis}
\end{figure*}

To be consistent with the 
input of ModelNet40, each input point cloud is randomly sampled to an average of 
2,048 points. The other experimental settings are consistent with previous 
experiments. 
Fig. \ref{kitti-vis} and Fig. \ref{3dmatch-vis} show visualization results of 
KITTI and 3DMatch, respectively.  
As shown in Table \ref{realworld}, we observe that
PCRDiffusion can still achieve stable results comparing other
methods. PCRDiffusion demonstrates remarkable generalization on real-world datasets, and this framework can still achieve optimal performance without the need for iterative processes.

%We observe that baselines are unstable in
%complex environments.
%We can summarize EGST 
%has best performance and is comfortable with real-world datasets. 

\subsection{Ablation Studies and Disscussion}
We conduct extensive ablation studies for a better understanding
of the various modules in PCRDiffusion framework. To achieve
a fair comparison, we utilize the same experimental settings
as Section \ref{setup} in ablation studies.

\noindent\textbf{Diffusion or Diffsuion-Free?}
In order to validate the potential significance of the diffusion model in point cloud registration tasks, we conduct a series of ablation experiments. Specifically, we exclude the transformation encoder. The experimental outcomes are presented in Table \ref{ab:diffusion}. We can observe that the model without diffusion has a clear performance drop, as they do not have  refinement property given by diffusion model.

In addition, we have uncovered another intriguing characteristic. Upon removing the transformation encoder, our two proposed simplified frameworks degrade to the original iterative-free PCRNet and RPMNet. Remarkably, under the same iterative-free conditions, the utilization of the diffusion model yields substantial performance enhancement. This presents a highly promising practical avenue: \textit{\textbf{integrating the diffusion model as a plug-and-play component into existing point cloud registration networks can yield exceptionally prominent results!}}
\begin{table}[h]
	\caption{Experimental results on model with or without diffusion.}
	\begin{center}
		\renewcommand\tabcolsep{4.5pt}
		\begin{tabular}{l|cccc}
			\toprule[1.5pt]
			%    \multicolumn{2}{c}{Part}                   \\
			%    \cmidrule(r){1-2}
			Method  & 
			\makecell{MAE($\textbf{R}$)} & \makecell{MAE($\textbf{t}$)} & 
			\makecell{ERROR($\textbf{R}$)}&\makecell{ERROR($\textbf{t}$)}\\
			\midrule
			(a) w/ diffusion (CF) &81.5854&0.4052&168.0197&0.7960\\ 
			(b) w/ diffusion (CB) &6.3867&0.0698&12.7412&0.1429\\ 
			\midrule
			(c) PCRDiffusion (CF) &1.6861&0.0216&3.2786&0.0448\\
			(d) PCRDiffusion (CB)&\textbf{0.7244}&\textbf{0.0153}&\textbf{1.2732}&\textbf{0.0318}\\
			\bottomrule[1.5pt]
		\end{tabular}
		\label{ab:diffusion}
	\end{center}
\end{table}

\noindent \textbf{Any Other Diffusion Objects?}
In our work, PCRDiffusion directly predicts object transformation from a random transformation. Furthermore, we unify the diffusion  transformation object $\textbf{G}$ to 7-dimensional vector (composed of a rotation quaternion and a translation). A natural idea arises: Is it possible to employ other transformation representation as diffusion object? We conduct experiments based on this notion, specifically, we replace the 7-dimensional vector into a 6-dimensional vector (composed of a rotation Euler angle and a translation), the distinction lies in the method of representing rotation. The experimental results are illustrated in Table \ref{ab:object}. We can observe utilizing Euler angles does not yield performance enhancements, primarily due to the significant disparity in values between Euler angles and translation vectors. This incongruence introduces singularities into the linear noise processes for diffusion model.
\begin{table}[h]
	\caption{Experimental results on different diffusion objects.}
	\begin{center}
		\renewcommand\tabcolsep{4.5pt}
		\begin{tabular}{l|cccc}
			\toprule[1.5pt]
			%    \multicolumn{2}{c}{Part}                   \\
			%    \cmidrule(r){1-2}
			Method  & 
			\makecell{MAE($\textbf{R}$)} & \makecell{MAE($\textbf{t}$)} & 
			\makecell{ERROR($\textbf{R}$)}&\makecell{ERROR($\textbf{t}$)}\\
			\midrule
			7-dimensional vectors  &\textbf{0.7244}&\textbf{0.0153}&\textbf{1.2732}&\textbf{0.0318}\\ 
			6-dimensional vectors &6.8240&0.0733&13.7542&0.1513\\ 
			\bottomrule[1.5pt]
		\end{tabular}
		\label{ab:object}
	\end{center}
\end{table}

\noindent \textbf{How to Fuse Noisy Transformation?}
During training, we construct  noisy transformation by adding Gaussion noise, and feed it to transformation estimation decoder after encoding. In our work, we fuse the encoded noisy transformation features $\mathcal{F}_{t}$ with the feature of source point cloud $\mathcal{F}_{\mathcal{P}}$. Additionally, we have also explored two alternative approaches: (A) concatenating all features and (B) fuse $\mathcal{F}_{t}$ with the feature of source point cloud $\mathcal{F}_{\mathcal{Q}}$. The experimental outcomes are illustrated in Table \ref{ab:fuse}. Practice proves that the model using approach A fails to converge and exhibits extreme instability. Furthermore, the fused approach demonstrates no significant variation in performance.
\begin{table}[h]
	\caption{Experimental results on different diffusion objects. $\oplus$ represents concatenation.}
	\begin{center}
		\renewcommand\tabcolsep{5.2pt}
		\begin{tabular}{l|cccc}
			\toprule[1.5pt]
			%    \multicolumn{2}{c}{Part}                   \\
			%    \cmidrule(r){1-2}
			Step  &  
			\makecell{MAE($\textbf{R}$)} & \makecell{MAE($\textbf{t}$)} & 
			\makecell{ERROR($\textbf{R}$)}&\makecell{ERROR($\textbf{t}$)}\\
			\midrule
			$\mathcal{F}_{t}\oplus\mathcal{F}_{\mathcal{P}}\oplus\mathcal{F}_{\mathcal{Q}}$ &-&-&-&-\\ 
			$\mathcal{F}_{\mathcal{P}}\oplus(\mathcal{F}_{t}+\mathcal{F}_{\mathcal{Q}})$&1.6887&0.0216&3.1686&0.0446\\
			$(\mathcal{F}_{t}+\mathcal{F}_{\mathcal{P}})\oplus\mathcal{F}_{\mathcal{Q}}$ &1.6861&0.0216&3.2786&0.0448\\ 
			\bottomrule[1.5pt]
		\end{tabular}
		\label{ab:fuse}
	\end{center}
\end{table}

\noindent \textbf{Accuracy vs. Speed.}
We test the inference speed of PCRDiffusion in Table \ref{accuracy-speed}. The run time is evaluated on an NVIDIA RTX 3090 GPU with a mini-batch size of 1. The experimental results show that utilizing correspondence-based PCRDiffusion (PCRDiffusion CB) and correspondence-free  PCRDiffusion (PCRDiffusion CF) respectively lead to the fastest inference speeds in all correspondence-based and correspondence-free methods. This is attributed to the iterative-free nature of our proposed framework, and it requires only a single step during reverse diffusion.

\begin{table}[h]
	\caption{Accuracy vs. Speed on ModelNet40.}
	\begin{center}
		\renewcommand\tabcolsep{2.8pt}
		\begin{tabular}{l|ccccc}
			\toprule[1.5pt]
			%    \multicolumn{2}{c}{Part}                   \\
			%    \cmidrule(r){1-2}
			Method  & 
			MAE($\textbf{R}$) & MAE($\textbf{t}$) & 
			\makecell{ERROR\\($\textbf{R}$)} &\makecell{ERROR\\($\textbf{t}$)}& Time(s)\\
			\midrule
			ICP ($\blacktriangle$) &11.7468&0.1686&23.1548&0.3497&0.0042\\ 
			PointNetLK ($\triangle$) &1.7362&0.4907&4.0421&0.9741&0.7111\\
			PCRNet ($\triangle$) &10.2498&0.1168&19.0487&0.2442&1.2318\\
			DeepGMR ($\blacktriangle$) &23.5162&0.3291&45.7786&0.6758&0.1053\\
			FGR ($\blacktriangle$) &1.9971&0.0268&4.3337&0.0582&0.0916\\
			DCP ($\blacktriangle$) &10.7846&0.1340&20.8744&0.2740&0.2819\\
			FMR ($\triangle$)  &1.5962&0.4886&3.8090&0.9751&1.0472\\
			RPMNet ($\blacktriangle$)  &1.4399&0.0170&2.9619&0.0357&0.1626\\
			\midrule
			PCRDiffusion (CF) ($\triangle$) &1.6861&0.0216&3.2786&0.0448&\textless $\text{10}^\text{-10}$\\
			PCRDiffusion (CB) ($\blacktriangle$)&\textbf{0.7244}&\textbf{0.0153}&\textbf{1.2732}&\textbf{0.0318}&0.0986\\
			\bottomrule[1.5pt]
		\end{tabular}
		\label{accuracy-speed}
	\end{center}
\end{table}

\noindent \textbf{Sampling Steps.}
We conduct experiments employing diverse sampling steps on correspondence-free method. The results are shown in Table \ref{ab:step}.
The results of these experiments indicate that PCRDiffusion yields optimal outcomes within a single step, and as the number of sampling iterations increases, performance rapidly converges to stability.
\begin{table}[h]
	\caption{Experimental results on different inference steps.}
	\begin{center}
		\renewcommand\tabcolsep{6pt}
		\begin{tabular}{c|cccc}
			\toprule[1.5pt]
			%    \multicolumn{2}{c}{Part}                   \\
			%    \cmidrule(r){1-2}
			Steps  &  
			\makecell{MAE($\textbf{R}$)} & \makecell{MAE($\textbf{t}$)} & 
			\makecell{ERROR($\textbf{R}$)}&\makecell{ERROR($\textbf{t}$)}\\
			\midrule
			1 &1.6861&0.0216&3.2786&0.0448\\ 
			2&1.6625&0.0215&3.1962&0.0445\\
			4 &1.6625&0.0215&3.1962&0.0445\\ 
			8&1.6625&0.0215&3.1962&0.0445\\
			\bottomrule[1.5pt]
		\end{tabular}
		\label{ab:step}
	\end{center}
\end{table}

\section{Conclusion}
\label{sec:conclusion}
In this work, we propose a novel point cloud registration paradigm,
PCRDiffsuion, by viewing object transformation as a denoising
diffusion process from noisy transformation to object transformation. Our
"noise-to-ground truth transformation" pipeline has several appealing advantage of Once-for-All: (i) We train the network only once, and no iteration strategy is required to update source point cloud during training and inference. (ii) We can adjust the number of denoising sampling steps to improve the registration accuracy or accelerate the inference speed. 
PCRDiffusion achieves favorable performance compared to
well-established point cloud registration methods.

To further explore the potential of diffusion model to solve point cloud registartion tasks, several future works are beneficial. An attempt is to apply PCRDiffusion to multimodel tasks, for example, utilizing the image as the condition to guide point cloud registration.

\bibliographystyle{IEEEtran}
\bibliography{references}

\begin{IEEEbiography}[{\includegraphics[width=1in,height=1.25in,clip,keepaspectratio]{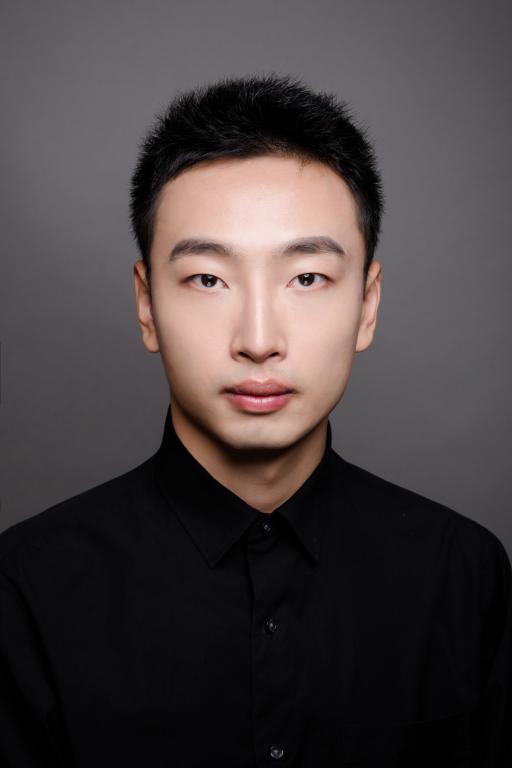}}]
	{Yue Wu} received the B.Eng. and Ph.D. degrees from Xidian University, Xi’an, China, in 2011 and 
	2016, respectively. 
	Since 2016, he has been a Teacher with Xidian University. He is currently an Associate Professor with 
	Xidian University. He has authored or co-authored more than 70 papers in refereed journals and 
	proceedings. His research interests include computational intelligence and its Applications. 
	He is the Secretary General of Chinese Association for Artificial Intelligence-Youth Branch, Chair of 
	CCF YOCSEF Xi’an, Senior Member of Chinese Computer Federation. He is Editorial Board Member for over 
	five journals, including Remote Sensing, Applied Sciences, Electronics, Mathematics.
\end{IEEEbiography}

\begin{IEEEbiography}[{\includegraphics[width=1in,height=1.25in,clip,keepaspectratio]{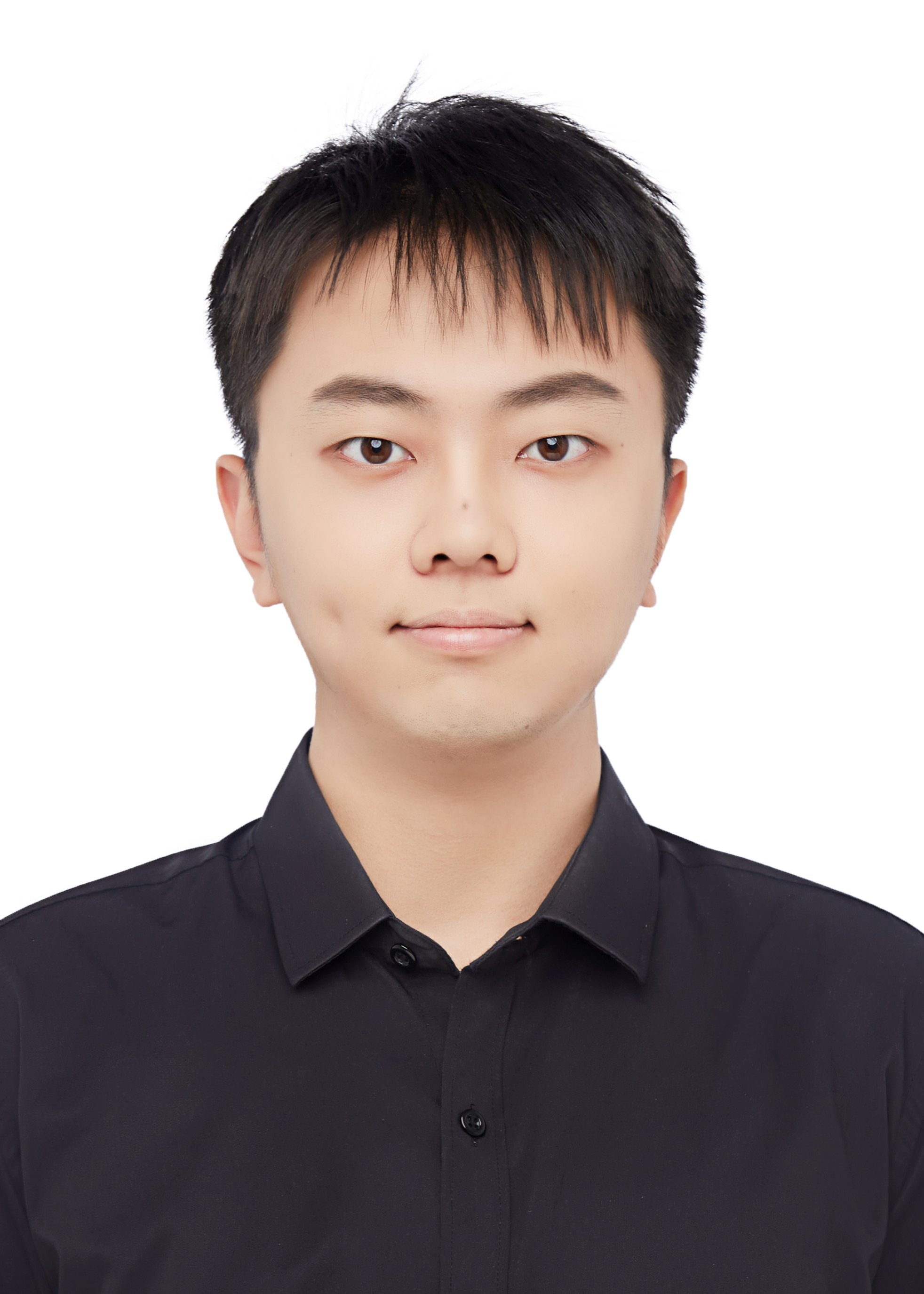}}]
	{Yongzhe Yuan}
	is currently pursuing the Ph.D
	degree with the School of Computer Science and
	Technology, the Key Laboratory of Collaborative Intelligence Systems of Ministry of Education, Xidian University, Xi’an.
	His research interests include deep learning, 3D computer vision and
	remote sensing image understanding.
\end{IEEEbiography}

\begin{IEEEbiography}[{\includegraphics[width=1in,height=1.25in,clip,keepaspectratio]{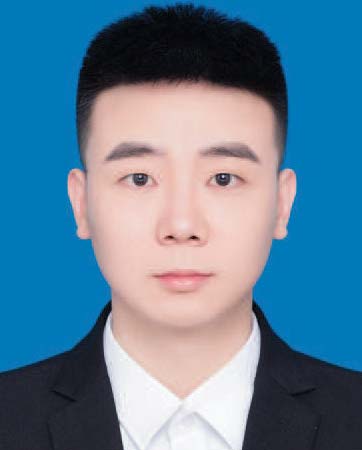}}]
	{Xiaolong Fan} 
	received the B.S. degree from Northwest
	University, Xi’an, China, in 2016. He is currently
	pursuing the Ph.D. degree in pattern recognition
	and intelligent systems with the School of Electronic
	Engineering, the Key Laboratory of Intelligent
	Perception and Image Understanding of Ministry of
	Education of China, Xidian University, Xi’an.
	His research interests include deep neural network,
	graph representation learning, and computer vision.
\end{IEEEbiography}

\begin{IEEEbiography}[{\includegraphics[width=1in,height=1.25in,clip,keepaspectratio]{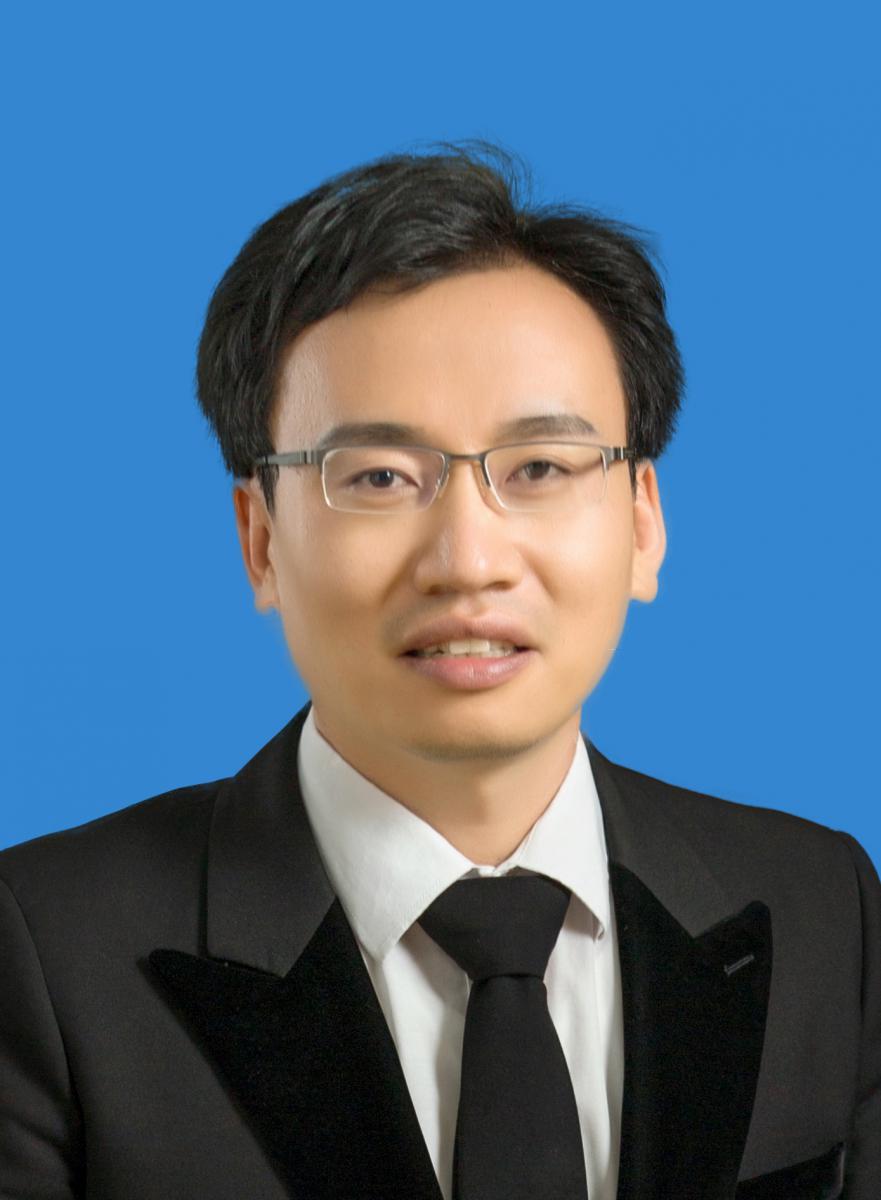}}]
	{Maoguo Gong} received the B.Eng. and Ph.D. degrees from Xidian University, 
	Xi’an, China, in 2003 and 2009, respectively. Since 2006, he has been a Teacher with Xidian 
	University. He was promoted to an Associate Professor and a Full Professor, in 2008 and 2010, 
	respectively, with exceptive admission. He has authored or coauthored over 100 articles in 
	journals and conferences. He holds over 20 granted patents as the first inventor. He is leading or 
	has completed over twenty projects as the Principle Investigator, funded by the National Natural 
	Science Foundation of China, the National Key Research and Development Program of China, and 
	others. His research interests are broadly in the area of computational intelligence, with 
	applications to optimization, learning, data mining, and image understanding. Prof. Gong is the 
	Executive Committee Member of Chinese Association for Artificial Intelligence and a Senior Member 
	of Chinese Computer Federation. He was the recipient of the prestigious National Program for 
	Support of the Leading Innovative Talents from the Central Organization Department of China, the 
	Leading Innovative Talent in the Science and Technology from the Ministry of Science and 
	Technology of China, the Excellent Young Scientist Foundation from the National Natural Science 
	Foundation of China, the New Century Excellent Talent from the Ministry of Education of China, and 
	the National Natural Science Award of China. He is an Associate Editor or an Editorial Board 
	Member for over five journals including the IEEE TRANSACTIONS ON EVOLUTIONARY COMPUTATION and the 
	IEEE TRANSACTIONS ON NEURAL NETWORKS AND LEARNING SYSTEMS.
\end{IEEEbiography}

\begin{IEEEbiography}[{\includegraphics[width=1in,height=1.25in,clip,keepaspectratio]{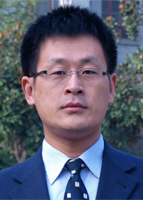}}]
	{Qiguang Miao} received
	the M.Eng. and Ph.D. degrees in computer science
	from Xidian University, Xi’an, China, in 1996 and
	2005, respectively.
	He is currently a Professor with the School of
	Computer Science and Technology, Xidian University. His research interests include intelligent 
	image processing and multiscale geometric representations
	for images.
\end{IEEEbiography}


%\vfill


\end{document}